\documentclass[final,3p,times,twocolumn]{elsarticle}

\usepackage{amsmath,amssymb,amsfonts}
\usepackage{algorithmic}
\usepackage{graphicx}
\usepackage{textcomp}
\usepackage{xcolor}
\usepackage{algorithm}
\usepackage{listings}
\usepackage{colortbl}
\usepackage{booktabs}
\usepackage{ulem}
\usepackage{multirow}
\usepackage{colortbl}
\usepackage{hyperref}

\usepackage{lineno}

\journal{Preprint}
\begin{document}
\begin{frontmatter}



\title{DSTSA-GCN: Advancing Skeleton-Based Gesture Recognition with Semantic-Aware Spatio-Temporal Topology Modeling}
\author[label2]{Hu Cui}
\author[label3]{Renjing Huang}
\author[label4]{Ruoyu Zhang}
\author[label2]{Tessai Hayama\corref{cor1}}
\ead{t-hayama@kjs.nagaokaut.ac.jp}
\cortext[cor1]{Corresponding author}
\affiliation[label2]{organization={Nagaoka University of Technology},
            addressline={1603-1 Kamitiomioka},
            city={Nagaoka-shi},
            postcode={9402188},
            state={Niigata},
            country={Japan}}
\affiliation[label3]{organization={Guizhou College of Electronic Science and Technology},
            addressline={Gui'an New District},
            city={Guiyang},
            postcode={550025},
            state={Guizhou},
            country={China}}
\affiliation[label4]{organization={Guizhou University},
            addressline={Huaxi District},
            city={Guiyang},
            postcode={550025},
            state={Guizhou},
            country={China}}            
\begin{abstract}
Graph convolutional networks (GCNs) have emerged as a powerful tool for skeleton-based action and gesture recognition, thanks to their ability to model spatial and temporal dependencies in skeleton data. However, existing GCN-based methods face critical limitations: (1) they lack effective \textit{spatio-temporal topology modeling} that captures dynamic variations in skeletal motion, and (2) they struggle to model \textit{multiscale structural relationships} beyond local joint connectivity.
To address these issues, we propose a novel framework called Dynamic Spatial-Temporal Semantic Awareness Graph Convolutional Network (DSTSA-GCN). DSTSA-GCN introduces three key modules: Group Channel-wise Graph Convolution (GC-GC), Group Temporal-wise Graph Convolution (GT-GC), and Multi-Scale Temporal Convolution (MS-TCN). GC-GC and GT-GC operate in parallel to independently model channel-specific and frame-specific correlations, enabling robust topology learning that accounts for temporal variations. Additionally, both modules employ a grouping strategy to adaptively capture multiscale structural relationships. Complementing this, MS-TCN enhances temporal modeling through group-wise temporal convolutions with diverse receptive fields.
Extensive experiments demonstrate that DSTSA-GCN significantly improves the topology modeling capabilities of GCNs, achieving state-of-the-art performance on benchmark datasets for gesture and action recognition, including SHREC’17 Track, DHG-14/28, NTU-RGB+D, and NTU-RGB+D-120. The code will be publicly available \url{https://hucui2022.github.io/dstsa_gcn/}.
\end{abstract}




\end{frontmatter}


\section{Introduction}
\label{introduction}
Gestures, as a distinct form of human action, can convey richer semantic information than general actions and even serve as an alternative to spoken language in communication. The human hand, with its intricate joint structure, enables a wide range of movements to express diverse meanings. In recent years, the advancement of depth sensors, known for their robustness in complex environments, has driven significant progress in skeleton-based gesture recognition. This technology is increasingly applied to areas such as sign language communication\cite{shen2024auslan, miah2024hand,desai2024asl}, robotic control\cite{qi2024computer,yu2024gesture}, virtual reality\cite{di2024towards}, and human-computer interaction\cite{modaberi2024role}.

Early methods for skeleton-based action and gesture recognition treated human joints as time-varying features or pseudo-images, using RNNs or CNNs to model spatial and temporal correlations. However, these approaches overlook the topological semantics of skeletal motion, leading to suboptimal performance. To address this gap, Yan et al. \cite{yan2018spatial} introduced ST-GCN, a framework that employs graph convolution (GCN) \cite{kipf2016semi} to capture skeletal motion patterns by representing joints as nodes and their natural connections as edges in a predefined graph. The GCN process is typically divided into three stages: \textit{feature extraction}, \textit{adjacency matrix construction}, and \textit{feature aggregation}. Among these, the \textit{construction of the adjacency matrix} is crucial, as it dictates how information flows between joints and bones in skeletal actions or gestures. ST-GCN \cite{yan2018spatial} manually defines the adjacency matrix based on natural joint connections, which restricts its ability to model relationships between non-naturally connected joints, limiting the representational capacity of GCNs.

\begin{figure*}[t]
	\centering
        \includegraphics[width=0.9\linewidth]{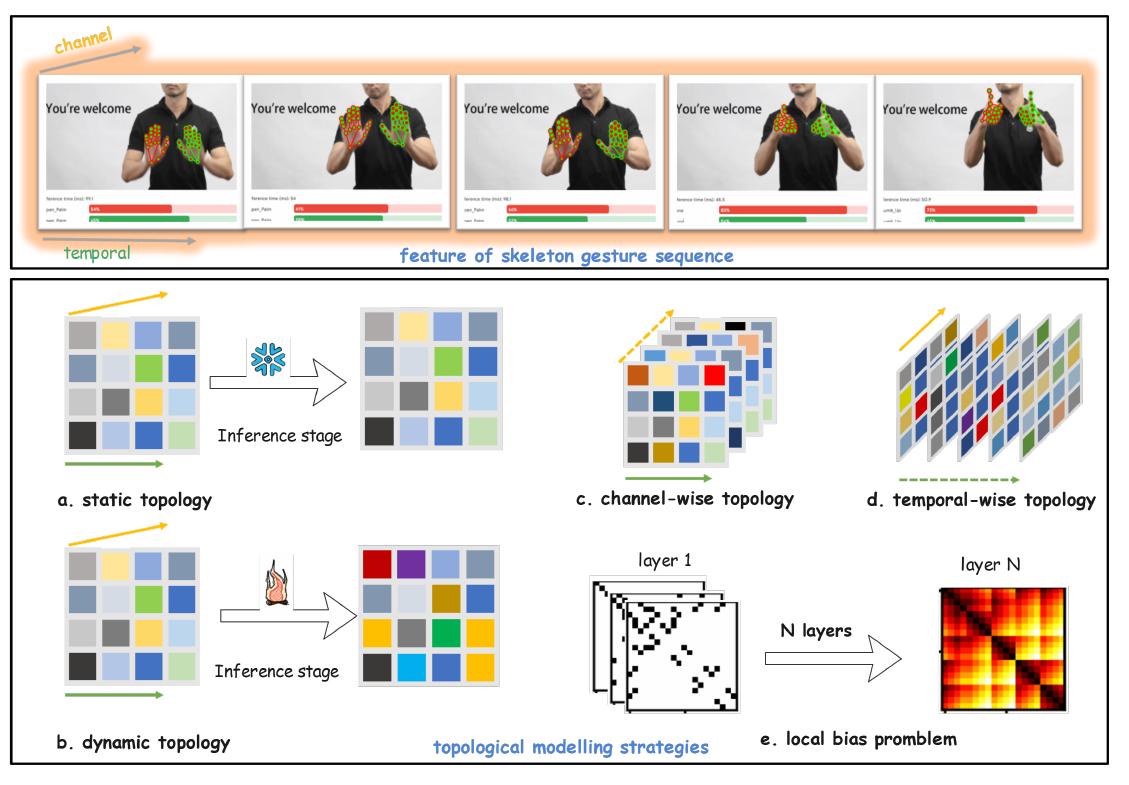}
	\caption{Illustration of the construction of different skeletal topologies. Solid arrows indicate shared topology in the channel or temporal dimension, while dashed lines indicate non-shared. Yellow arrows indicate the direction of the channel and green arrows indicate the direction of the temporal dimension. \textbf{c} and \textbf{d} is based on the dimension-specific of \textbf{b}. \textbf{a} is parameterized in training stage and does not vary in the inference phase with the samples.}
	\label{fig:1}
\end{figure*}
To improve GCN expression flexibility, following approaches \cite{shi2019two,li2019actional,ye2020dynamic, zhang2020semantics,cheng2020skeleton} propose to add a dynamic topology graph which depends on the input samples to fine-tune the manually predefined graph (Fig.\ref{fig:1}a+b). The introduction of dynamic graphs significantly improves the topology modeling capability of GCNs. However, these methods still force feature aggregation to follow the same topological pattern across all channels. Since different channels capture distinct motion characteristics, employing a \textit{shared topology} may not be optimal. 
In response, \cite{cheng2020decoupling, chen2021channel,duan2022dg} introduced channel-specific dynamic topologies to aggregate features across different channels.
Unlike \textit{topology-shared} methods \cite{shi2019two,li2019actional,ye2020dynamic, zhang2020semantics,cheng2020skeleton}, these \textit{topology-non-shared} approaches \cite{cheng2020decoupling, chen2021channel,duan2022dg} use different topologies (Fig.\ref{fig:1}c) for feature extraction and aggregation on each channel, allowing the model to better capture expressive interaction information.

However, these dynamic topology-non-shared GCNs overlook two important aspects of skeleton gesture recognition by primarily focusing on spatial topological representation. \textbf{(1) Temporal properties in action topology modeling.} These methods perform non-shared dynamic fine-tuning for adjacency matrixes in the channel dimension, but these adjacency matrixes are still shared in the temporal dimension, which limits the flexibility of GCN for modeling temporal information. Intuitively, the inter-joint interactions and responses should vary at different time steps of a movement. This is especially true for gestures, where the complexity and dexterity of the hand joints and bones allow for the expression of diverse semantics. For example, the gesture of ``you are welcome", palm outward thrusts in the first half and thumbs up in the second half obviously hold different topological semantics, as shown in Fig. \ref{fig:1}. \textbf{(2) Multiscale relationship beyond local joints.} Previous methods generally achieve higher order responses between joints by stacking the GCN layers as shown in top of Fig.\ref{fig:1}e. \label{qusetion:2}
This leads to the model only focusing on local information in the shallow layers, and becoming biased towards local information in deeper layers. However, in reality, interactions between distant joints are equally or more important as those in close range. For example, the gesture of ``pinch (with one's fingers)",  the primary semantics derive from the topologically distant fingertips.

To address these limitations, we propose the Dynamic Spatial-Temporal Semantic Awareness Graph Convolutional Network (DSTSA-GCN). The core of DSTSA-GCN consists of three key modules: Group Channel-wise Graph Convolution (GC-GC), Group Temporal-wise Graph Convolution (GT-GC), and Multi-Scale Temporal Convolution (MS-TCN). GC-GC and GT-GC are responsible for modeling the spatial features in terms of channel-specific and temporal-specific characteristics, respectively. Meanwhile, MS-TCN is designed to capture the temporal dynamics of the input data.

The topology graph in GC-GC consists of a dynamic part which is channel-specific (Fig.\ref{fig:1}c) and a static part which is group-specific. 
In order to adaptively capture interactions at multiple scales, we set the static part as learnable parameters. 
Since the static graphs in each layer are generated by random initialization (or relative distance matrix), there is no local bias problem.
GT-GC models temporal-wise topologies (Fig.\ref{fig:1}d) in the same way. 
In order to improve the computational efficiency, we compress the temporal dimension when modeling the channel-wise topologies, and compress the channel dimension when modeling the temporal-wise topologies, the fused features of GC-GC and GT-GC not only refine the spatial topology modeling of gesture skeletons, but also enhance the awareness of temporal topology variations. 
For temporal modeling, we extend the traditional fixed-kernel temporal convolution to a multiscale temporal convolutional layer (MS-TCN) to enhance the model's sensitivity to action velocity and lower the computational costs.
We conduct extensive experiments in skeleton-based gesture recognition and whole-body action recognition  and  compare  our  results  with  competitive  baselines on two popular gesture bench-mark datasets: SHREC’17 Track and DHG-14/28 and two action bench-mark datasets: NTU-RGB and NTU-RGB 120. The experiments show that our model achieves state-of-the-art in terms of accuracy.
Our main contributions can be summarized as follows:
\begin{itemize}
    \item  We propose dynamic, non-shared graph convolutions for both channels (GC-GC) and temporal dimensions (GT-GC) to capture finer spatial-temporal topological features in skeleton-based gesture actions.
    \item We incorporate multi-scale strategies into both spatial and temporal modeling to address the spatial bias of GCNs and the insensitivity of TCNs.
    \item Extensive experiments on the SHREC'17 Track, DHG-14/28 gesture datasets and NTU-RGB, NTU-RGB 120 action datasets demonstrate that our DSTSA-GCN can more effectively capture spatio-temporal dependencies in the topology.
\end{itemize}

\begin{figure*}[t]
	\centering
		\centering
		\includegraphics[width=\linewidth]{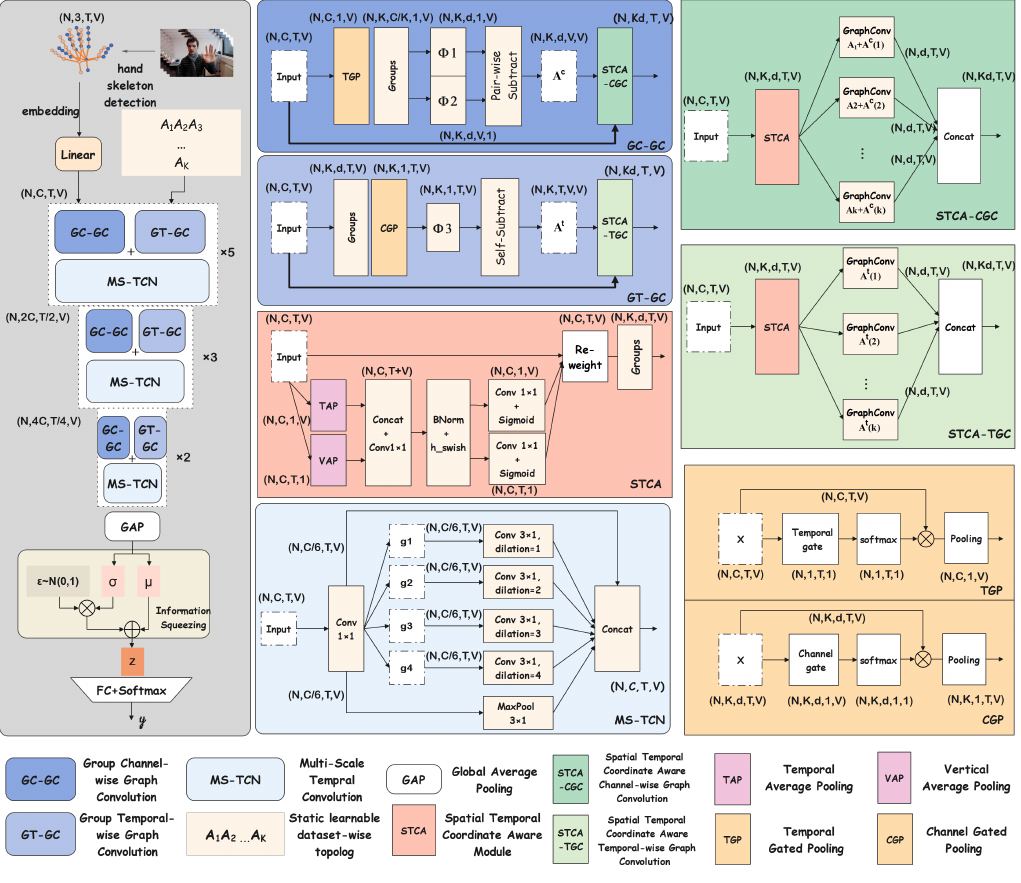}
	\caption{The architecture of Dynamic Spatial-Temporal Semantic Awareness Graph Convolutional Network.}
	\label{fig:2}
\end{figure*}
\section{Related Work}
\subsection{Graph Neural Networks}
Graph neural networks (GNNs) \cite{velivckovic2023everything, xu2018powerful, wu2020comprehensive, zhu2020graph, liu2024towards, zheng2024structure, zheng2024gnnevaluator} have proven to be as effective for processing non-Euclidean data, such as graphs, as convolutional neural networks (CNNs) are for handling Euclidean data, such as images. 
There are two series that garner the most attention in recent years, namely, spectral GNNs and the spatial GNNs. 
Spectral GNNs\cite{wang2022powerful,bo2023survey} leverage the Laplacian spectrum to define graph convolution, applying learned filters to graph signals in the Fourier domain. On the other hand, spatial GNNs \cite{danel2020spatial, pei2020geom} perform graph convolutions in the spatial domain by updating node features through sampling and aggregating features from local neighbors in the graph.
Among the many GNN variants, Graph Convolutional Networks (GCNs) \cite{kipf2016semi} introduced a simplified first-order approximation of ChebNet for local spectral convolution. Similar to spatial GNNs, GCNs update node features in three steps: (1) transforming input features into high-level representations, (2) constructing the adjacency matrix, and (3) aggregating features based on the adjacency matrix.
GCNs' mathematical interpretability and formal simplicity have made them a common baseline for subsequent research on graph-structured data.\label{relatedwork_gnn}

\subsection{Skeleton-Based Action and Gesture Recognition}
Earlier approaches to skeleton-based action and gesture recognition treated the human skeleton as either time-varying sequences or pseudo-images, focusing on hand-crafted features for classification \cite{de2016skeleton, lai2018cnn+, de2019heterogeneous}. However, these methods overlooked the semantic interactions between joints. Recent advancements address this limitation by constructing spatio-temporal graphs, where joints are represented as vertices and bones as edges, and leveraging graph neural networks (GNNs) to model joint relationships \cite{yan2018spatial,shi2019two,li2019actional,ye2020dynamic,zhang2020semantics,cheng2020skeleton,shi2019two,li2019actional,ye2020dynamic, zhang2020semantics,cheng2020skeleton}. These developments highlight the critical role of interaction semantics in understanding skeletal actions.

ST-GCN \cite{yan2018spatial} revolutionarily represents the naturally connected human skeleton as a graph and utilizes graph convolutional networks (GCNs) combined with temporal convolutional networks (TCNs) to extract spatial topological and motion features.
To enhance the flexibility and expressiveness of ST-GCN, AS-GCN \cite{li2019actional} introduces actional and structural links to extend the static, predefined graph, effectively capturing implicit joint correlations through an actional-structural graph convolutional network.
2s-AGCN \cite{shi2019two} and SGN \cite{zhang2020semantics} dynamically model the correlation between two joints using self-attention mechanism. Dynamic GCN \cite{ye2020dynamic} takes full account of the contextual information of each joint from a global perspective to learn correlations between any pairs of joints. These methods have stronger generalization ability due to \textit{dynamic} topologies, but still aggregate features in  different channels with the same topology as ST-GCN which limits the model performance. To overcome this limitation, DC-GCN \cite{cheng2020decoupling} split channels into decoupling groups, and each channel group has a dynamic adjacent matrix. CTR-GCN \cite{chen2021channel} proposes to model channel-wise topologies through learning a shared topology and refining it with channel-specific dynamic adjacent matrix for each channel.   Inspired by these works, we propose temporal-wise topologies to further enhance the GCNs ability to dynamic actions.
\subsection{Nolocal-Range Dependencies in GCN}
In addition to enhancing the flexibility of ST-GCNs by introducing \textit{dynamic, non-shared topologies} in GCNs, another effective approach is improving the model's ability to capture long-range dependencies within skeleton graphs. MST-GCN \cite{chen2021multi} enriches the spatial receptive field by incorporating residual-connected multi-scale spatial graph convolutions. Similarly, MSTGNN \cite{feng2021multi} employs a fine-to-coarse three-scale graph structure to capture multi-scale information effectively. In this paper, we design both graph convolution and temporal convolution with a multi-grouping strategy, enabling the model to capture multi-scale interaction information across both spatial and temporal dimensions.

\section{Method}
In this section, we begin by defining the related notations and formulating conventional graph convolution. Next, we present the overall architecture of our proposed framework, the Dynamic Spatial-Temporal Semantic Awareness Graph Convolutional Network (DSTSA-GCN). Following this, we provide a detailed explanation of the core components of our model: Grouped Channel-wise Graph Convolution (GC-GC), Grouped Temporal-wise Graph Convolution (GT-GC), and Multi-Scale Temporal Convolution (MS-TCN).
\subsection{Preliminaries}\label{sec:preliminary}
\noindent \textbf{Notations.} A gesture skeleton can be represented as a graph $\mathbf{G} = \{\mathcal{V,E}\}$ with joints as vertices $\mathcal{V} = (v_1,v_2,...,v_N)$ and bones  as edges $\mathcal{E}$. For 3D skeleton, the joint $v_i = (x_i, y_i, z_i)$ where $x_i, y_i, z_i$ are the coordinates of the i-th joints. We use the adjacency matrix $\mathbf{A} \in \mathbb{R}^{V \times V} $ to represent $\mathcal{E}$ and the element $\mathbf{A}_{i,j}$ reflects the dependency between $v_i$ and $v_j$. The features of vertices $\mathcal{V}$ are represented as $\mathcal{X} \in \mathbb{R}^{C\times T \times V }$, in where $C$ is the number of channels, $V$ is the number of joints in one frame and $T$ is the number of frames. In our experiments, we use four modalities  of skeleton data. Given the \textit{joint data} of joint modality $\mathbf{v}_{t,i} = (x_{t,i}, y_{t,i}, z_{t,i})$ and $\mathbf{v}_{t,j} = (x_{t,j}, y_{t,j}, z_{t,j})$, the \textit{bone data} of bone modality is $\mathbf{e}_{t,ij} = (x_{t,i} - x_{t,j}, y_{t,i} - y_{t,j}, z_{t,i} - z_{t,j})$, the \textit{joint motion} data is defined as $\mathbf{jm}_{ti} = \mathbf{v}_{t,i} - \mathbf{v}_{t-1,i}$, and the \textit{bone motion} data is defined as $\mathbf{bm}_{t,ij} = \mathbf{e}_{t,ij} - \mathbf{e}_{t-1, ij}$.

\noindent \textbf{Spatial Graph Convolution.} 
The GCN-based action and gesture recognition model consists of several spatio-temporal GCN blocks, and \textit{spatial graph convolution} is the key component.
$\mathbf{X} = \mathcal{X}_{:,t,:} \in \mathbb{R}^{C \times V}$ denotes the spatial features of skeleton gesture.
The \textit{spatial graph convolution} can be expressed as :
\begin{equation}
    {S\_GCN}(\mathbf{X})=\sum_{p \in \mathcal{P}} \widetilde{\mathbf{A}}^{(p)} \mathbf{X^\top} \mathbf{W}^{(p)}
    \label{eq_stgcn}
\end{equation}
where $\mathcal{P} = \{root, centripetal, centrifugal\}$ denotes the partition subsets in ST-GCN \cite{yan2018spatial}. 
$\mathbf{W}^{(p)} \in \mathbb{R}^{C \times C'}$ is trainable parameters, and $\mathbf{A}^{(p)}$ is normalized adjacency matrix of the graph of subset $p$. 
\begin{equation}
    \widetilde{\mathbf{A}}^{(p)} = \mathbf{D}^{(p)^{-\frac{1}{2}}} \mathbf{A}^{(p)} \mathbf{D}^{(p)^{-\frac{1}{2}}}
\end{equation}
where $\mathbf{D}_{i i}^{(p)}=\sum_j\left(\mathbf{A}_{i j}^{(p)}\right)+\varepsilon$ and $\varepsilon$ is a small constant for numerical stability.
It is worth noting that $\mathcal{P}$ can be represented in a variety of different ways \cite{chen2021channel, duan2022dg,shi2019two,li2019actional,liu2023temporal}.
As we said in Sec.\ref{introduction}, the GCN-based methods can be divided into static and dynamic methods depending on whether the topology is dynamically adjusted during the inference process and topology-sharing and topology-non-sharing methods depending on whether the topology is shared among different channels. For  static  methods, $\mathbf{A}_{i j}^{(p)}$ is predefined or trained and fixed during the inference process, for dynamic methods, $\mathbf{A}_{i j}^{(p)}$ is generated depending on the input sample. For topology-sharing methods, $\mathbf{A}_{i j}^{(p)}$ is shared for all channels, and $\mathbf{A}^{(p)} \in \mathbb{R}^{V \times V}$, for toplogy-non-sharing methods, $\mathbf{A}_{i j}^{(p)}$ is specific for each channel, and $\mathbf{A}^{(p)} \in \mathbb{R}^{C \times V \times V}$.

\subsection{Dynamic Spatial-Temporal Semantic Awareness Graph Convolutional Network}\label{DSTSA-GCN)}
As shown in Fig. \ref{fig:2}, our model is constructed by three-stage stacking of the basic modules consisting of Group Channel-wise Graph Convolution (GC-GC), Group Temporal-wise Graph Convolution (GT-GC) and Multi-Scale Temporal Convolution (MS-TCN). $\{5, 3, 2\}$ stacks are performed in the \{1st, 2nd, 3rd\} stage, followed by one downsample layer at the end of \{1st, 2nd\} stage. 
The model takes a 3D skeleton gesture sequence $\mathcal{X}_0 \in \mathbb{R}^{3 \times T \times V}$ as input, and then proceeds to map each joint to a $C$-dimensional feature $\mathcal{X} \in \mathbb{R}^{C \times T \times V}$. 
Suppose the input of the $L_{th}$ block is $\mathcal{X}_{\ell-1}$, after spatial modeling and temporal modeling, the output can be expressed as:
\begin{equation}
    \mathcal{X}_{\ell} = TC(GC(\mathcal{X}_{\ell-1})) + \mathcal{X}_{\ell-1}
\end{equation}
where GC denotes spatial graph convolution and TC denotes temporal convolution.
Specifically, $GC(\cdot)$ consists of two components, GC-GC and GT-GC.
\begin{equation}
    GC(\cdot) = \xi(GC\text{-}GC(\cdot), GT\text{-}GC(\cdot))
\end{equation}
where $\xi$ is a fusion function of GC-GC and GT-GC. When fused in parallel, it can be expressed as: 
\begin{equation}
    \xi(GC\text{-}GC(\cdot), GT\text{-}GC(\cdot)) = a  \cdot GC\text{-}GC(\cdot) + b  \cdot  GT\text{-}GC(\cdot)
\end{equation}
where $a, b$ represent two learnable parameters, which will be optimized during the training process.
To capture kinematic features of different lengths and variations, we apply a multi-scale grouped temporal convolution in place of traditional temporal convolution for temporal modeling.
It is important to note that while we evaluated the impact of incorporating information bottleneck (IB) \cite{hjelm2018learning,alemi2016deep,chi2022infogcn}loss on final performance, but the majority of the experiments were conducted using cross-entropy (CE) loss.


\subsection{Grouped Channel-wise Graph Convolution}\label{GCGCN}
The GC-GC module adopts a \textit{channel-wise topology-non-sharing} strategy, as described in Sec.~\ref{introduction} and Sec.~\ref{relatedwork_gnn}. Similar to other GCN methods, GC-GC operates in three stages: feature transformation, topology modeling, and feature aggregation. The process of channel-wise graph convolution (C-GC) can be formally expressed as:
\begin{equation}
    C\text{-}GC(\mathcal{X}) = \mathcal{A}(\mathcal{F}(\mathcal{X}), \mathcal{T}(\mathcal{D}(\mathcal{X}), \mathbf{A}) )
    \label{eq:gcg}
\end{equation}
where $\mathcal{A}(\cdot)$ represents the \textbf{a}ggregation function, $\mathcal{F}(\cdot)$ is the \textbf{f}eature transformation function, and $\mathcal{T}(\cdot)$ is the \textbf{t}opology modeling function. Here, $\mathcal{D}(\cdot)$ represents the dynamic topology component, while $\mathbf{A}$ corresponds to the static topology component.

Notably, when $\mathcal{F}(\mathcal{X}) = \mathcal{X} \mathbf{W}^\top$ and $\mathcal{T}(\mathbf{A}) = \mathbf{D}^{-\frac{1}{2}} \mathbf{A} \mathbf{D}^{-\frac{1}{2}}$, the C-GC operation simplifies to the equation used in ST-GCN (Eq.~\ref{eq_stgcn}).

\noindent \textbf{Feature Transformation.}  
As illustrated in the red and green blocks in Fig.~\ref{fig:2}, unlike conventional methods that employ a simple linear transformation for topology-shared graph convolution, we leverage a Spatio-Temporal Coordinate-Aware (STCA) module to enrich the transformed features with spatio-temporal locational information. The STCA module can be intuitively understood as a dynamic position encoding or position attention mechanism, expressed as:
\begin{align}
    \mathcal{\hat{X}} &= \mathcal{F}(\mathcal{X}) \\
    &= STCA(\mathcal{X})\mathbf{W}^\top
    \label{eq:fx}
\end{align}
where $\mathcal{\hat{X}}$ represents the transformed features.

The STCA module captures intra-frame and temporal position information while incorporating joint spatio-temporal coordinate data, which is crucial for dynamic gesture actions. The process is divided into two steps: spatio-temporal joining $\mathcal{J}(\cdot)$ and decoupling. The spatio-temporal joining step is defined as:
\begin{equation}
    \mathcal{J}(\mathcal{X}) = \delta(F_1([TAP(\mathcal{X}), VAP(\mathcal{X})]))  \label{eq-9}
\end{equation}
where $TAP(\mathcal{X}) = \frac{1}{T} \sum_{0 \leq i < T} \mathcal{X}(:, i, :)$ and $VAP(\mathcal{X}) = \frac{1}{V} \sum_{0 \leq j < V} \mathcal{X}(:, :, j)$ represent average pooling along the temporal and joint dimensions, respectively. Here, $[\cdot,\cdot]$ denotes concatenation along the last dimension, $F_1(\cdot)$ is a $1 \times 1$ convolutional transformation, and $\delta(\cdot)$ is a non-linear activation function (e.g., the hardswish function). The resulting tensor $\mathcal{J}(\mathcal{X}) \in \mathbb{R}^{C/r \times (T+V)}$ jointly encodes spatial and temporal information, where $r$ is a reduction ratio for controlling module size.

Next, $\mathcal{J}(\mathcal{X})$ is split along the last dimension into two tensors, which are processed using two separate $1 \times 1$ convolutions $f_t(\cdot)$ and $f_v(\cdot)$ to restore the channel dimensions:
\begin{equation}
    \mathbf{g}^t = \sigma(f_t(\mathcal{J}(\mathcal{X})_{:,:T}))
\end{equation}
\begin{equation}
    \mathbf{g}^v = \sigma(f_v(\mathcal{J}(\mathcal{X})_{:,V:}))
\end{equation}
where $\sigma(\cdot)$ is the sigmoid function. The resulting tensors $\mathbf{g}^t \in \mathbb{R}^{C \times T \times 1}$ and $\mathbf{g}^v \in \mathbb{R}^{C \times 1 \times V}$ are expanded to serve as temporal and spatial weights, respectively.

Finally, the STCA module is expressed as:
\begin{equation}
    STCA(\mathcal{X}) = \mathbf{g}^v \cdot \mathbf{g}^t \cdot \mathcal{X}
\end{equation}
where $\cdot$ denotes element-wise multiplication. This process enables the STCA module to effectively encode and enrich the spatio-temporal features of input data.

\noindent \textbf{Channel-wise Topology Modeling.}  
As illustrated in the dark blue and dark green blocks in Fig.~\ref{fig:2}, the channel-wise topology modeling utilizes the adjacency matrix $\mathbf{A} \in \mathbb{R}^{V \times V}$ as a shared, static topology graph for all channels (for brevity, we omit the grouping step here). This adjacency matrix is learned through backpropagation and fixed during the inference stage, meaning it represents the topology at the dataset level. In contrast, a dynamic topological graph $\mathbf{A}^c = \mathcal{D}(\mathcal{X})$ captures topological information at the sample level. This dynamic topology is defined as:
\begin{equation}
    \mathcal{D}(\mathcal{X}) = \mathcal{M}_p(\Phi_1(TGP(\mathcal{X})), \Phi_2(TGP(\mathcal{X})))
\end{equation}
where $TGP(\cdot)$ represents temporal-gated pooling, defined as:
\begin{equation}
    TGP(\mathcal{X}) = \sum_{t}^{T} \mathcal{X} \cdot \text{softmax}\left( \frac{1}{CV} \sum_{c}^{C} \sum_{v}^{V} X_{c, :, v} \right)
\end{equation}
and $\Phi_1$, $\Phi_2$ are $1 \times 1$ convolutional transformation functions. 

The function $\mathcal{M}_p(\cdot)$ computes the distances between $\Phi_1(x_i)$ and $\Phi_2(x_j)$ along the channel dimension, utilizing non-linear transformations of these distances as the channel-wise topological relationship between nodes $v_i$ and $v_j$:
\begin{equation}
    \mathcal{M}_p(\Phi_1(x_i), \Phi_2(x_j)) = \theta(\Phi_1(x_i) - \Phi_2(x_j))
\end{equation} \label{eq-thata1}
where $\theta(\cdot)$ is an activation function (e.g., tanh or sigmoid).

Finally, the channel-wise topology modeling function $\mathcal{T}(\cdot)$ is expressed as:
\begin{align}
    \mathbf{\hat{A}} &= \mathcal{T}(\mathcal{D}(\mathcal{X}), \mathbf{A}) \label{eq:16}\\
    &= \mathcal{D}(\mathcal{X}) \cdot \alpha + \mathbf{A} \\
    &= \mathbf{A}^c \cdot \alpha + \mathbf{A},
\end{align}
where $\alpha$ is a trainable scalar. The addition is performed in a broadcast manner, where $\mathbf{A} \in \mathbb{R}^{V \times V}$ is added to each channel of $\mathbf{A}^c \in \mathbb{R}^{C \times V \times V}$ (grouping is omitted for brevity).

\noindent \textbf{Grouped Feature Aggregation.}
By substituting Eq.~\ref{eq:fx} and Eq.~\ref{eq:16} into Eq.~\ref{eq:gcg}, we obtain:
\begin{equation}
    C\text{-}GC(\mathcal{X}) = \mathcal{A}(\mathcal{\hat{X}}, \mathbf{\hat{A}} )
    \label{eq:gcn2}
\end{equation}
Since feature aggregation is performed on each channel graph, this can be expressed in matrix form as:
\begin{equation}
    \mathcal{A}(\mathcal{\hat{X}}, \mathbf{\hat{A}}) = \left[ \mathcal{\hat{X}}_{1,:,:} \mathbf{\hat{A}}_{1,:,:} \ \| \ 
    \mathcal{\hat{X}}_{2,:,:} \mathbf{\hat{A}}_{2,:,:} \ \| \ \dots \ \| \ 
    \mathcal{\hat{X}}_{C,:,:} \mathbf{\hat{A}}_{C,:,:} \right]
\end{equation}
where $\|$ denotes the concatenation operation along the channel dimension.

As shown in Eq.~\ref{eq_stgcn}, previous feature aggregation methods~\cite{chen2021channel, duan2022dg, shi2019two, li2019actional, liu2023temporal} often design multiple static topology graphs and fuse the resulting topological features to extract more expressive interaction semantics. However, this results in a larger number of parameters, proportional to the partition subsets $\mathcal{P}$ as seen in Eq.~\ref{eq_stgcn}. 

Inspired by the design of grouped convolutions, as depicted in the dark green block in Fig.~\ref{fig:2}, we propose a channel-grouped graph convolution for $\mathcal{X} \in \mathbb{R}^{C \times T \times V} \to \mathbb{R}^{K \times \frac{C}{K} \times T \times V}$. In this approach, the static graphs $\mathbf{A} \in \mathbb{R}^{K \times V \times V}$ are shared within each group, while the dynamic graphs remain channel-specific.

Our grouped channel-wise graph convolution can then be expressed as:
\begin{align}
    GC\text{-}GC(\mathcal{X}) &= \left[ C\text{-}GC_{1}(\mathcal{X}) \ \| \ C\text{-}GC_{2}(\mathcal{X}) \ \| \ \cdots \ \| \ C\text{-}GC_{K}(\mathcal{X}) \right] \\
    &= \left[ \mathcal{A}(\mathcal{\hat{X}}_{:\frac{C}{K},:,:}, \mathbf{\hat{A}}_{1}) \ \| \ \mathcal{A}(\mathcal{\hat{X}}_{\frac{C}{K}:\frac{2C}{K},:,:}, \mathbf{\hat{A}}_{2}) \ \| \ \cdots \right]
    \label{eq-23}
\end{align}\label{eq-k1}
where $\hat{A}_{i} = \mathbf{A}^c\left[\frac{(i-1)C}{K}:\frac{iC}{K}\right] \cdot \alpha + \mathbf{A}[i]$.

Compared to previous methods, the number of parameters in our approach is independent of the number of groups, which significantly enhances the multi-scale topology modeling capability while maintaining the same complexity constraints.

\subsection{Grouped Temporal-wise Graph Convolution}\label{GTGCN}

The temporal-wise graph convolution (T-GC) process within the GT-GC module can be expressed similarly to C-GC as follows:
\begin{equation}
    T\text{-}GC(\mathcal{X}) = \mathcal{\bar{A}}(\mathcal{\bar{F}}(\mathcal{X}), \mathcal{\bar{T}}(\mathcal{\bar{D}}(\mathcal{X})))
    \label{eq:24}
\end{equation}

\noindent \textbf{Feature Transformation.}
The feature transformation process $\mathcal{\bar{F}}(\cdot)$ is analogous to the one described in Eq.~\ref{eq:fx}. To reduce the number of model parameters, we set it as a shared module, as shown in the light green and dark green blocks in Fig.~\ref{fig:2}. Specifically, it is given by:
\begin{equation}
    \mathcal{\bar{F}}(\mathcal{X}) = \mathcal{F}(\mathcal{X})
    \label{eq-25}
\end{equation}


\noindent \textbf{Temporal-wise Topology Modeling.} 
Unlike the channel-wise topology modeling process described in Sec.~\ref{GCGCN}, temporal-wise topology modeling involves only the dynamic graph component $\mathbf{A}^t = \mathcal{\bar{D}}(\mathcal{X})$. This process can be expressed as:
\begin{equation}
    \mathcal{\bar{D}}(\mathcal{X}) = \mathcal{M}_s(\Phi_3(CGP(\mathcal{X})))
\end{equation}
where $CGP(\cdot)$ denotes the temporal-gated pooling, as illustrated in the yellow block in Fig.~\ref{fig:2}, and can be expressed as:
\begin{equation}
    CGP(\mathcal{X}) = \sum_{c}^{C} \mathcal{X} \cdot \text{softmax}\left( \frac{1}{TV} \sum_{t}^{T} \sum_{v}^{V} \mathcal{X}_{:, t, v} \right)
\end{equation}
Here, $\Phi_3(\cdot)$ denotes a $1 \times 1$ convolutional transformation function, and $\mathcal{M}_s(\cdot)$ calculates self-pairwise distances between $\Phi_3(x_i)$ and $\Phi_3(x_j)$ along the temporal dimension. The non-linear transformation of these distances is used as the temporal-wise topological relationship between $v_i$ and $v_j$, expressed as:
\begin{equation}
    \mathcal{M}_s(\Phi_3(\cdot)) = \theta(\Phi_3(x_i) - \Phi_3(x_j))
\end{equation} \label{eq-theta2}
Finally, the temporal-wise topology modeling function $\mathcal{\bar{T}}(\cdot)$ is given by:
\begin{align}
\mathbf{A}^t &= \mathcal{\bar{T}}(\mathcal{\bar{D}}(\mathcal{X})) \\
 &= \mathcal{\bar{D}}(\mathcal{X})
\label{eq:30}
\end{align}


\noindent \textbf{Grouped Feature Aggregation.}
Bringing Eq.~\ref{eq-25} and Eq.~\ref{eq:30} into Eq.~\ref{eq:24} gives:
\begin{equation}
    T\text{-}GC(\mathcal{X}) = \mathcal{\bar{A}}(\mathcal{\hat{X}}, \mathbf{A}^t)
    \label{eq:31}
\end{equation}
Feature aggregation is performed on each temporal graph as:
\begin{equation}
    \mathcal{\bar{A}}(\mathcal{\hat{X}}, \mathbf{A^{t}}) = [\mathcal{\hat{X}}_{:,1,:} \mathbf{{A^{t}}}_{1,:,:} ||
    \mathcal{\hat{X}}_{:,2,:} \mathbf{{A^{t}}}_{2,:,:} ||
    \dots||
    \mathcal{\hat{X}}_{:,T,:} \mathbf{{A^{t}}}_{T,:,:}
    ]
\end{equation}
Similar to grouped channel-wise graph convolution, grouped temporal-wise graph convolution can be expressed as:
\begin{align}
GT\text{-}GC(\mathcal{X}) &= [T\text{-}GC_{1}(\mathcal{X})||T\text{-}GC_{2}(\mathcal{X})||\cdots|| T\text{-}GC_{K}(\mathcal{X})]\\
&= [\mathcal{\bar{A}}(\mathcal{\hat{X}}_{:\frac{C}{K} ,:,:}, \mathbf{A^{t}_{1}})||\mathcal{\bar{A}}(\mathcal{\hat{X}}_{\frac{C}{K}:\frac{2C}{K} ,:,:}, \mathbf{A^{t}_{2}})||\cdots)]
\label{eq-34}
\end{align}\label{eq-k2}
By comparing Eq.~\ref{eq-23} and Eq.~\ref{eq-34}, it can be observed that the GC-GC models topological information from the channel-wise perspective, whereas the GT-GC models topological information from the temporal-wise perspective. The combination of these two approaches enables the model to capture richer spatio-temporal topological information.


\subsection{Multi-Scale Temporal Convolution}\label{MSTCN}
To model actions with different durations, we design a multi-scale temporal modeling module (MS-TCN) following previous works \cite{chen2021channel,liu2020disentangling,liu2023temporal,cui2024joint,cui2024stsd,duan2022dg}. As shown in the bright-blue block in Fig.~\ref{fig:2}, the MS-TCN has a branching structure similar to group convolution. Each branch contains a $1 \times 1$ convolution to reduce the channel dimension. The first four branches consist of four temporal convolutions with different dilation rates $\{1,2,3,4\}$, enabling the model to capture temporal dependencies at multiple scales. 

The remaining two branches include a MaxPool operation and a shortcut connection, both preceded by a $1 \times 1$ convolution. We conduct ablation experiments on the multi-branch structure and select the best combination as the final temporal convolution module of our MS-TCN.

\section{Experiments}
To verify the generality of the model, we use DHG-14/28 \cite{de2016skeleton} and SHREC’17 Track \cite{de20173d} for gesture recognition and NTU-RGB+D \cite{shahroudy2016ntu} and NTU-RGB+D 120 \cite{liu2019ntu} for human action recognition. We first performed an exhaustive ablation study on the SHREC17 dataset to validate the effectiveness of the proposed model components. Then, we evaluate our model on all four datasets and compare it with state-of-the-art methods.
\subsection{Datasets}
\textbf{SHREC :} The SHREC’17 Track dataset comprises 2800 gesture sequences performed by 28 participants, with each gesture repeated between 1 and 10 times. Gestures are performed in two distinct manners: using a single finger or the entire hand. Gesture sequence lengths vary between 20 and 50 frames. Each sequence is labeled according to either 14 or 28 gesture classes, depending on the number of fingers used and the gesture type. The dataset is divided into 1960 sequences for training and 840 sequences for testing, following the evaluation protocol in\cite{de20173d},\cite{song2022dynamic} and \cite{shi2020decoupled}. Recognition accuracy can be computed based on either the 14-class or 28-class labeling scheme.
The dataset is specifically designed for the SHREC’17 Track competition, held in conjunction with the Eurographics 3DOR 2017 Workshop.

\textbf{DHG :} The DHG-14/28 dataset contains 2800 video sequences of 14 hand gestures, each performed five times by 20 participants, using either one finger or the whole hand. Similar to SHREC’17 Track, there are two classification benchmarks: 14 gestures for coarse classification and 28 gestures for fine-grained analysis. The 3D coordinates of 22 hand joints are captured using an Intel RealSense camera. A leave-one-subject-out cross-validation strategy is used, with data from 19 participants for training and the remaining participant for testing. This process is repeated 20 times, and the average accuracy is reported as the final result.

\textbf{NTU-60 :}
The NTU-RGB+D (NTU-60) dataset is one of the most widely used benchmarks for indoor-captured 3D action recognition. It comprises 56,880 skeleton sequences categorized into 60 action classes, performed by 40 subjects. The data is captured using three Kinect V2 cameras from different viewpoints, with each skeleton sequence providing 25 joints per subject. The dataset supports two evaluation protocols: (1) Cross-Subject (CS), where 20 subjects are used for training and the other 20 for testing, and (2) Cross-View (CV), where training data comes from two camera views (0° and 45°), while testing data comes from a third view (-45°).

\textbf{NTU-120 :}
The NTU-RGB+D 120 (NTU-120) dataset is an extended version of NTU-60 for 3D action recognition. It consists of 114,480 skeleton sequences spanning 120 action classes, performed by 106 participants across 32 distinct camera setups. The dataset includes two evaluation protocols: Cross-Subject (CS), where half of the participants are used for training and the remaining half for testing, and Cross-Setup (CE), where sequences from setups with odd IDs are used for training, and those from setups with even IDs are reserved for testing.

\begin{table*}[ht]
    \centering
    \begin{minipage}[t]{0.45\textwidth}
            \begin{minipage}[t]{\textwidth}
                \centering
                \caption{Effectiveness of each component on SHREC'17 in jiont mode.} \label{ex-ab1}
                    \resizebox{\textwidth}{!}{
                        \begin{tabular}{lllll} \hline
                            \textbf{Method}         & \textbf{Param}. & \textbf{FLOPs}. & \textbf{14Gestures}(\%) & \textbf{28Gestures}(\%)  \\
                            \hline\hline
                            w/o GC-GC             &1.55M    &1.53G    &95.71$_{\downarrow 0.96}$            &92.06$_{\downarrow 1.87}$                 \\
                            w/ GC-GC w/o TGP  &1.99M    &1.79G    &96.12$_{\downarrow 0.55}$               &92.97$_{\downarrow 0.96}$                 \\
                            w/ GC-GC w/o STCA &1.89M    &1.78G    &95.95$_{\downarrow 0.72}$                &92.61$_{\downarrow 1.33}$                 \\
                            \hline
                            w/o GT-GC         &1.77M    &1.79G    &95.83$_{\downarrow 0.84}$                &93.05$_{\downarrow 0.88}$                 \\
                            w/ GT-GC w/o CGP  &1.99M   &1.79G     &96.07$_{\downarrow 0.60}$                &93.27$_{\downarrow 0.66}$                 \\
                            w/ GT-GC w/o STCA &1.89M    &1.78G    &96.34$_{\downarrow 0.33}$                &93.45$_{\downarrow 0.48}$                 \\
                            \hline
                            TCN            &2.68M   &2.93G     &95.60$_{\downarrow 1.07}$                &92.87$_{\downarrow 1.06}$                 \\
                            MS-TCN         &1.99M   &1.79G     &\textbf{96.67}                &\textbf{93.93}                  \\
                            \hline
                        \end{tabular}
                    }
            \end{minipage}
            \vfill
            \vspace{18pt}
            \begin{minipage}[t]{\textwidth}
                \centering
                \caption{Configurations exploration of GC-GC and GT-GC  on SHREC'17 in jiont mode. ${\triangledown}$ denotes comparison with the results of $\theta = Tanh, K = 4$, ${\downarrow}$ denotes comparison with $\theta =Tanh, K=8$. }\label{ex-ab2}
                \resizebox{\textwidth}{!}{
                    \begin{tabular}{llllll}
                        \hline
                        \multicolumn{1}{l}{\textbf{Method}} & \textbf{K} & $\mathbf{\theta}$  & \textbf{Param}. & \textbf{14Gesture}(\%) & \textbf{28Gesture}(\%)  \\ \hline\hline
                        A                           & 3      & Tanh    &1.96M        &96.43               &92.97                \\
                        B                           & 4      & Tanh    &1.97M        &\uline{\textbf{96.67}}${\triangledown}$   &\uline{93.93}${\triangledown}$                \\
                        C                           & 8      & Tanh    &1.99M        &\textbf{96.67}${\downarrow}$               &\textbf{94.17}${\downarrow}$                \\
                        D                           & 12     & Tanh    &1.97M        &96.04               &92.98                \\
                        E                           & 8      & Relu    &1.99M        &95.98$_{\downarrow 0.67}$               &92.85$_{\downarrow 1.32}$                \\
                        F                           & 4     & Relu    &1.97M        &95.21$_{\triangledown 1.46} $              &93.16$_{\triangledown 0.77}$                \\
                        G                           & 4      & Sigmoid &1.97M        &95.07$_{\triangledown1.60} $               &92.67$_{\triangledown 1.26}$                \\
                        H                           & 8     & Sigmoid &1.99M        &95.32$_{\downarrow 1.35}$               &92.74$_{\downarrow 1.43}$                \\
                        I                           & 8      & Softmax &1.99M        &96.59$_{\downarrow 0.08}$               &93.66$_{\downarrow0.51}$                \\
                        J                           & 4     & Softmax &1.97M        &96.38$_{\triangledown0.29} $               &93.54$_{\triangledown 0.39}$               \\ \hline
                    \end{tabular}
                    \label{tab:left_table}
                }
            \end{minipage}
    \end{minipage}
    \hfill
    \begin{minipage}[t]{0.45\textwidth}
            \centering
            \begin{minipage}[t]{0.8\textwidth}
                \centering  
                \caption{Comparison of the performance of differently constructed STCA modules. $\Uparrow$ denotes parallel and $\Rightarrow$ denotes series.}\label{ex-ab3}
                \resizebox{\textwidth}{!}{
                    \begin{tabular}{lllll}
                        \hline
                        \textbf{Method}  & \textbf{Param}. & \textbf{Flops}. & \textbf{14Gestures}(\%) & \textbf{28Gestures}(\%)  \\ \hline\hline
                        w/o any. &1.78M        &1.89G        &95.28                &92.83                 \\
                        SCA     &1.96M        &1.79G        &95.77$_{\uparrow 0.49}$                &93.22$_{\uparrow 0.39}$                 \\
                        TCA     &1.96M        &1.79G        &96.02$_{\uparrow 0.74}$                &93.09$_{\uparrow 0.26}$                 \\
                        SCA $\Uparrow$ TCA &1.99M       &1.79G      &96.21$_{\uparrow 0.93}$        &93.47$_{\uparrow 0.64}$          \\
                        SCA$\Rightarrow$TCA &1.99M      &1.79G     & 96.39$_{\uparrow 1.11}$       &  93.54$_{\uparrow 0.71}$            \\
                        STCA    &1.99M        &1.79G        &\textbf{96.67}$_{\uparrow \textbf{1.39}}$                &\textbf{94.17}$_{\uparrow \textbf{1.34}}$           \\  \hline
                    \end{tabular}
                }
            \end{minipage}
            \vfill
            \begin{minipage}[t]{0.9\textwidth}
                \centering  
                \caption{Performance of MS-TCN in different configurations. M denotes the Maxpooling branch in light blue block of Fig. \ref{fig:2}. S denotes the shortcut branch, $g_i$ denotes the other branches with different dilations.}\label{ex-ab4}
                \resizebox{\textwidth}{!}{
                    \begin{tabular}{clllll}
                        \hline
                        \textbf{Method} & \textbf{Branches} & \textbf{Param}. & \textbf{Flops}. & \textbf{14Gestures}(\%) & \textbf{28Gestures}(\%)  \\ \hline\hline
                        TCN(d5)         & -                & 2.68M           & 2.93G           &95.70                    &92.93                     \\
                        A               & g1               & 2.69M           &2.55G                 &95.21                    &92.44                      \\
                        B               & g1,g2            & 2.30M           & 2.13G                &95.84                    &93.57                      \\
                        C               & g1,g2,g3         & 2.16M            &1.99G                 &96.49                    &93.88                      \\
                        D               & g1,g2,g3,g4      & 2.10M           & 1.92G                &96.62                    &94.11                          \\
                        E               & g1,g2,g3,g4,g5      & 2.04M           & 1.87G                &96.14                    &93.77                          \\
                        F               & M,g1,g2,g3,g4    & 2.03M             & 1.85G                &96.60                    &94.14                      \\
                        G               & S,g1,g2,g3,g4    & 2.03M            & 1.83G                &96.53                    &94.01                         \\
                        H               & M,S,g1,g2,g3,g4  & 1.99M             & 1.79G                &\textbf{96.67 }                   &\textbf{94.17}   \\
                        I               & M,S,g1,g2,g3,g4,g5 &1.95M             & 1.74G                & 96.44                 & 93.79
                        
                        \\ \hline
                    \end{tabular}
                }
            \end{minipage}
            \vfill
            \begin{minipage}[t]{0.7\textwidth}
                \centering
                \caption{Effectiveness of static topology and initialization strategies.}  \label{ex-ab5}
                \resizebox{\textwidth}{!}{
                    \begin{tabular}{llll}\hline
                        \textbf{Method}          &  \textbf{14Gestures}(\%) & \textbf{28Gestures}(\%)  \\ \hline\hline
                        w/o A           &        95.30                 &93.12                  \\
                        Spatial 1. w/ A &        96.19$_{\uparrow 0.89}$                  &93.93$_{\uparrow 0.80}$                   \\
                        Spatial 2. w/ A &         96.54$_{\uparrow 1.24}$                 &94.01$_{\uparrow 0.89}$                   \\
                        Rand. w/ A      &        96.67$_{\uparrow 1.37}$                  &\textbf{94.17}$_{\uparrow \textbf{1.05}}$                   \\
                        Dis. w/ A       &        \textbf{96.81}$_{\uparrow \textbf{1.51}}$                  &93.82$_{\uparrow 0.70}$                  
                        \\ \hline
                        \end{tabular}
                }
            \end{minipage}
    \end{minipage}
\end{table*}

\subsection{Training Details}
All experiments are conducted on the Pytorch platform with one RTX A6000 GPU card.
To show the generalization of our methods, we use the same base channel $C=64$.
For SHREC’17 Track and DHG-14/28, the input gesture sequence is randomly/uniformly sampled to 150 frames for training/test splits. 
Batch size is 64. 
For NTU-60 and NTU-120, the input action sequence is randomly/uniformly sampled to 64 frames for training/test splits. 
Batch size is 32. 
We use the SGD optimizer with the initial learning rate of 0.1, weight decay of 0.0004, and Nesterov momentum of 0.9. The learning rate is divided by 10 in 70 and 100 epochs.
The training is ended in 170 epochs. The warm up epoch for learning rate is 20.
Cross entropy loss (CE) for classiﬁcation are used to train the networks without any data enhancement strategies.

\begin{figure*}[t]
	\centering
		\centering
		\includegraphics[width=\linewidth]{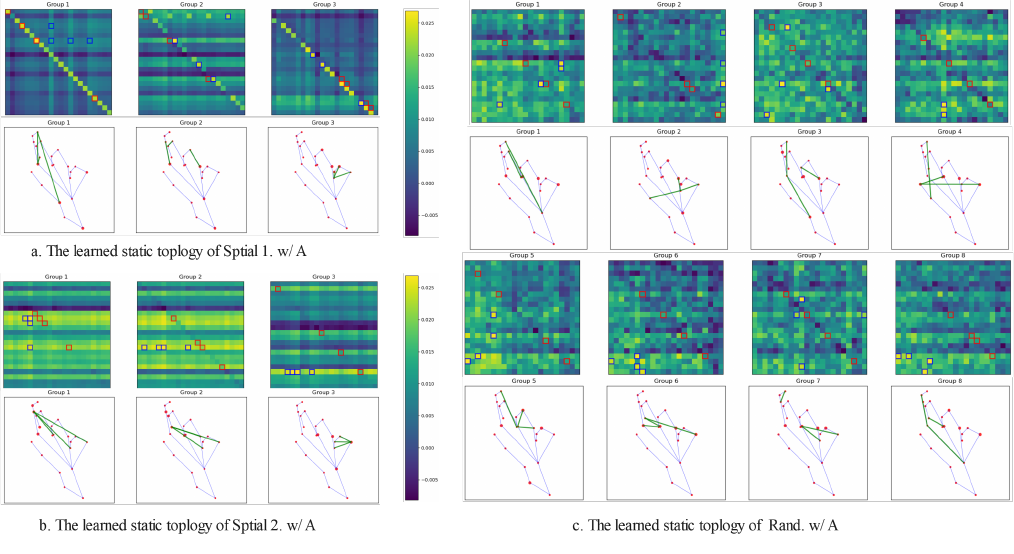}
	\caption{Visualization of the static topology graphs learned from different initialization strategies. Blue boxes indicate the maximum four distant interactions, and red indicates the maximum four self interactions.}
	\label{fig:3}
\end{figure*}

\begin{table*}[t]
    \centering
    \begin{minipage}[t]{\textwidth}
        \centering
        \caption{Recognition results of 20 different subjects as the validation set of the DHG-14/28 Dataset. } \label{ex-dhg datasets}
            \resizebox{\textwidth}{!}{
                \begin{tabular}{llllllllllllllllllllll}
                \hline
                \multicolumn{2}{l}{Val. Participant}       & 1     & 2     & 3     & 4     & 5     & 6     & 7     & 8     & 9     & 10    & 11    & 12    & 13    & 14    & 15     & 16    & 17    & 18    & 19   & 20  \\ \hline\hline
                \multirow{5}{*}{14Gesture.} & J. Acc.(\%)  & 90.72 & 81.43 & 98.57 & 95.71 & 90.00 & 86.48 & 90.71 & 94.28 & 89.28 & 95.71 & 93.57 & 90.00 & 90.00 & 87.14 & 97.14  & 96.43 & 90.71 & 91.43 & 97.86 & 90.71\\
                                            & JM. Acc.(\%) & 87.86 & 84.29 & 97.14 & 94.29 & 89.29 & 83.57 & 87.14 & 90.71 & 89.29 & 92.14 & 95.00 & 90.00 & 88.57 & 86.28 & 96.43  & 95.71 & 90.00 & 89.29 & 97.14 & 92.14\\
                                            & B. Acc.(\%)  & 90.71 & 78.57 & 90.71 & 87.14 & 90.71 & 86.43 & 87.86 & 92.86 & 87.14 & 89.28 & 94.28 & 89.26 & 77.14 & 84.29 & 86.43  & 90.00 & 87.14 & 86.43 & 92.85 &82.86\\
                                            & BM. Acc.(\%) & 88.57 & 80.71 & 95.00 & 88.57 & 87.86 & 82.14 & 85.71 & 91.43 & 87.86 & 90.00 & 94.29 & 88.57 & 75.00 & 86.43 & 89.29  & 92.14 & 87.57 & 90.71 & 92.14 & 86.43\\
                                            & Acc.(\%)     & 94.28 & 90.00 & 99.29 & 97.86 & 95.00 & 92.14 & 93.57 & 95.71 & 96.43 & 97.86 & 95.71 & 92.86 & 91.43 & 91.43 & 100.00 & 97.14 & 94.29 & 92.86 & 99.29 & 93.57\\ \hline
                \multirow{5}{*}{28Gesture}  & J. Acc.(\%)  & 89.29 & 76.43 & 97.14 & 90.71 & 92.85 & 83.57 & 89.29 & 90.71 & 89.28 & 92.14 & 93.57 & 92.85 & 86.43 & 86.64 & 94.28  & 94.28 & 90.71 & 85.71 & 96.43 &89.29 \\
                                            & JM. Acc.(\%) & 90.00 & 77.14 & 96.48 & 91.42 & 90.00 & 85.00 & 84.28 & 85.71 & 85.00 & 91.43 & 95.72 & 89.29 & 83.57 & 84.29 & 95.00  & 94.28 & 90.71 & 88.57 & 94.29 & 88.57\\
                                            & B. Acc.(\%)  & 89.28 & 75.71 & 88.57 & 85.71 & 91.43 & 85.00 & 85.00 & 90.00 & 85.00 & 86.43 & 92.14 & 88.57 & 77.86 & 85.71 & 85.00  & 97.14 & 87.14 & 87.86 & 92.14 & 82.14\\
                                            & BM. Acc.(\%) & 90.00 & 77.14 & 87.14 & 82.14 & 90.71 & 84.29 & 77.86 & 94.28 & 83.57 & 88.57 & 94.28 & 89.29 & 75.71 & 83.57 & 90.00  & 89.29 & 85.00 & 86.43 & 93.57 & 80.71\\
                                            & Acc.(\%)     & 92.86 & 82.86 & 98.57 & 91.43 & 94.29 & 92.14 & 92.86 & 96.43 & 95.00 & 94.29 & 95.71 & 94.29 & 87.86 & 90.71 & 98.57  & 95.71 & 95.00 & 92.14 & 97.86 & 92.86\\ \hline
                \end{tabular}
            }
    \end{minipage}
    \vfill
    \begin{minipage}[t]{0.48\textwidth}
        \centering
        \caption{Accuracy comparison with state-of-the-art methods on SHREC’17 Track and DHG-14/28 datasets.} \label{ex:gesture_datasets}
        \resizebox{\textwidth}{!}{
            \begin{tabular}{llllll} \hline
            \multirow{2}{*}{Method} & \multirow{2}{*}{Publisher} & \multicolumn{2}{c}{SHREC}  & \multicolumn{2}{c}{DHG}        \\ \cline{3-6}
                       &             & 14Gesture (\%) & 28Gesture(\%) & 14Gesture(\%) & 28Gesture(\%)  \\ \hline\hline
            HG-GCN\cite{li2019spatial}     & JIVP2019            & 92.8           & 88.3      & 89.2          & 85.3           \\
            ST-GCN\cite{yan2018spatial}     & AAAI2018           & 92.7           & 87.7      & 91.2          & 87.1           \\
            Shift-GCN\cite{cheng2020skeleton}  & CVPR2020            & 95.5           & 89.4      & 93.2          & 87.4           \\
            HPEV\cite{liu2020decoupled}       & CVPR2020            & 94.9           & 92.3      & 92.5          & 88.9           \\
            STA-GCN\cite{han2024spatio}    & VC2020            & 95.4           & 91.8      & 91.5          & 87.7           \\
            ResGCNeXt\cite{peng2023efficient}  & TCDS2024            & 95.36          & 93.1      & -             & -              \\
            CTR-GCN\cite{chen2021channel}    & ICCV2021            & 96.1           & 94.4      & 93.1          & 90.5           \\
            TD-GCN\cite{liu2023temporal}     & TMM2023            & 97.02          & 95.36     & 93.9          & 91.4           \\
            STDA-GCN\cite{han2024spatio}   & Ele.2024       & \uline{97.14}     & \textbf{95.84}     & \uline{94.2}          & \uline{92.1}           \\ \hline
            DSTSA-GCN(J.)    &                & {96.67}    & {94.17}     & -         & -          \\
            DSTSA-GCN(JM.)    &                & {95.52}    & {91.19}     & -         & -          \\
            DSTSA-GCN(B.)    &                & {89.40}    & {87.38}     & -         & -          \\
            DSTSA-GCN(BM.)    &                & {89.05}    & {86.07}     & -         & -          \\
            DSTSA-GCN    &                & \textbf{97.74}    & \uline{95.37}     & \textbf{95.04}         & \textbf{93.57}          \\ \hline
            \end{tabular}
        }
    \end{minipage}
    \hfill
    \begin{minipage}[t]{0.48\textwidth}
        \centering
        \caption{Accuracy comparison with state-of-the-art methods on NTU-RGB+D and NTU-RGB+D 120 datasets.}\label{ex:action datasets}
        \resizebox{\textwidth}{!}{
            \begin{tabular}{llllll} \hline
                \multirow{2}{*}{Method} & \multirow{2}{*}{Publisher} & \multicolumn{2}{c}{NTU-60} & \multicolumn{2}{c}{NTU-120}  \\ \cline{3-6}
                                        &                            & C-Subject(\%) & C-View(\%)  & C-Subject(\%) &  C-Setup(\%)              \\ \hline\hline
                ST-GCN\cite{yan2018spatial}                  & AAAI2018                     & 81.5    & 88.0             & 89.2   & 85.3                \\
                SGN\cite{zhang2020semantics}                     & CVPR2020                   & 89.0    & 94.5             & 79.2   & 81.5                \\
                2s-AGCN\cite{shi2019two}                 & CVPR2019                   & 88.5    & 95.1             & 82.9   & 84.9                \\
                Shift-GCN\cite{cheng2020skeleton}               & CVPR2020                   & 90.7    & 96.5             & 85.9   & 87.6                \\
                Dynamic GCN\cite{ye2020dynamic}             & MM2020                     & 91.5    & 96.0             & 87.3   & 88.6                \\
                CTR-GCN\cite{chen2021channel}                 & ICCV2021                     & 92.4    & 96.8             & 88.9   & 90.6                \\
                SAN-GCN\cite{tian2023skeleton}                 & TMM2023                      & 92.1    & 96.2             & 88.7   & 90.1                \\
                BlockGCN\cite{10658569}                & CVPR2024             & \textbf{93.1}~ ~ & \uline{97.0} & \textbf{90.3}   & \textbf{91.5}                \\
                LG-SGNet\cite{wu2025local}                & PR2025               & \textbf{93.1}    & 96.7~ ~      & \uline{89.4}   & \uline{91.0}                \\ \hline
                DSTSA-GCN(J.)    &                & {90.13}    & {95.48}     & {85.42}         & {87.37}          \\
                DSTSA-GCN(JM.)    &                & {88.18}    & {93.27}     & {81.58}         & {84.90}          \\
                DSTSA-GCN(B.)    &                & {90.13}    & {95.74}     & {86.80}         & {88.92}          \\
                DSTSA-GCN(BM.)    &                & {87.94}    & {93.12}     & {82.33}         & {84.15}          \\
                DSTSA-GCN                 &                       & \uline{92.78}   & \textbf{97.03}  & 89.12   & 90.97                   \\ \hline
                \end{tabular}
                }
    \end{minipage}
\end{table*}

\subsection{Ablation Studies}
In this section, we conduct ablation studies to analyze the key designs of our DSTSA-GCN. (1) We first analyze the role of each component of DSTSA-GCN. (2) We then explore different conﬁgurations of GC-GC and GT-GC. (3) After that, we explore the construction of the STCA module. (4) Finally we explore the different structures and effects of the MS-TCN module. 
\\

\noindent\textbf{The Effectiveness of Each Component.}
The contributions of the model's components are evaluated in Tab. \ref{ex-ab1}. Notably, the GC-GC module has the most significant impact on performance. Disabling this module results in an accuracy drop of 0.96\% and 1.89\% for the 14gesture and 28gesture criteria, respectively. This decline is attributed to the loss of channel-wise topology modeling and the complete absence of data-level static topology modeling. Conversely, disabling the GT-GC module only reduces accuracy by 0.84\% and 0.88\%, as it only removes temporal-wise topology modeling while preserving static topology modeling through the GC-GC module.

Replacing temporal-gated pooling (TGP) in GC-GC with average pooling reduces performance by 0.55\% and 0.96\%, indicating that different frames in an action sequence contribute unevenly to channel topology. Similarly, channel-gated pooling (CGP) plays a comparable role in channel compression within the GT-GC module.

Disabling the STCA in GC-GC causes accuracy to drop by 0.72\% and 1.33\%, whereas disabling it in GT-GC leads to smaller declines of 0.33\% and 0.48\%. This suggests that channel dimensions are inherently more complex than temporal dimensions, making spatio-temporal semantic information more influential in channel modeling. Finally, the MS-TCN module not only reduces model complexity compared to TCN but also enhances performance by effectively capturing actions of varying durations.

\noindent\textbf{Configurations Exploration of GC-GC and GT-GC.}
As shown in Tab. \ref{ex-ab2}, we investigated the impact of the number of groups $K$ (in Eqs. \ref{eq-k1}, \ref{eq-k2}) and the choice of the distance activation function $\theta$ (in Eqs. \ref{eq-thata1}, \ref{eq-theta2}), which is shared between the GC-GC and GT-GC modules.

Initially, with $\theta$ fixed to Tanh, we tested different values of $K$. When $K$ was set to 4 and 8, the results achieved the second-best (96.7\% on 14Gesture, 93.93\% on 28Gesture) and the best performance (96.67\% on 14Gesture, 94.17\% on 28Gesture), respectively. Subsequently, we replaced Tanh with other activation functions, including ReLU, Sigmoid, and Softmax. Among these, only Softmax produced comparable results to Tanh, with slight performance drops of -0.08\% on 14Gesture and -0.51\% on 28Gesture when $K=8$, and -0.29\% on 14Gesture and -0.39\% on 28Gesture when $K=12$. 

We attribute this behavior to the non-negative outputs of Sigmoid and ReLU, which limit their ability to effectively capture correlations in the data.


\noindent\textbf{Construction of STCA.}
As discussed in Sec. \ref{GCGCN}, Sec. \ref{GTGCN}, and shown in Tab. \ref{ex-ab1}, the STCA module serves as the core of feature transformation for GC-GC and GT-GC, enabling them to capture richer semantics related to the spatio-temporal locations of joints.

In Tab. \ref{ex-ab3}, we use the results with the STCA module disabled as a baseline. When the interactions between spatial and temporal dimensions are decoupled—i.e., by disabling the concatenation operation in Eq. \ref{eq-9}—we obtain the spatial coordinate-aware (SCA) and temporal coordinate-aware (TCA) modules, respectively. The results show that in the decoupled state, whether SCA and TCA are connected in series or in parallel, the performance is inferior compared to the spatio-temporal coupled state.

This indicates that dynamic actions should be treated as a unified spatio-temporal entity. Even a single frame inherently carries implicit dynamic temporal properties, emphasizing the importance of considering both spatial and temporal dimensions together for effective modeling.


\noindent\textbf{Effectiveness of MS-TCN.} 
As shown in Tab. \ref{ex-ab4}, employing only two branches in B (dilation = 1, 2) achieves accuracy comparable to TCN while significantly reducing model complexity. Optimal performance is observed when the number of branches is expanded to 4, though the optimal configuration may vary depending on the dataset and base channel size. Additionally, the inclusion of shortcut and temporal max-pooling branches enhances training stability and provides slight improvements in final performance.

\noindent\textbf{Effectiveness of the Static Topology and Initialization Strategies.} 
In Eq. \ref{eq:gcg}, $\mathbf{A}$ represents the static topology component, which is parameterized during training and fixed during inference. As detailed in Tab. \ref{ex-ab5}, we evaluate several configurations for this component. \textit{Spatial 1} refers to the static topology matrix generated when $\mathcal{P} = \{self, in, out\}$, while \textit{Spatial 2} is generated using $\mathcal{P} = \{root, centripetal, centrifugal\}$, following \cite{yan2018spatial,chen2021channel,liu2023temporal} and Eq. \ref{eq_stgcn}. Both configurations consist of three groups ($K=3$). The static matrices in \textit{Spatial 1} are initialized using self-connections, in-degree matrices, and out-degree matrices, while those in \textit{Spatial 2} are initialized based on topological distances of three neighboring joints relative to the wrist (gesture datasets) or the middle of the spine (full-body action datasets).

As shown in Tab. \ref{ex-ab5}, both \textit{Spatial 1} and \textit{Spatial 2} improve static topological representation at the dataset level. However, their fixed number of groups (or subsets, as described in \cite{yan2018spatial,chen2021channel,liu2023temporal}) results in a local bias problem, which limits the ability to capture multi-scale relationships, as discussed in Sec. \ref{qusetion:2} and illustrated in Fig. \ref{fig:1}. 

In Fig. \ref{fig:3}, the learned topology graphs of the network's last layer are visualized. In Fig. \ref{fig:3}a, the static topology values are clearly biased toward neighboring joints and themselves (evidenced by the near-diagonal overlap of the blue box), making it difficult to capture distant joint interactions. Similarly, Fig. \ref{fig:3}b shows that distant interactions are restricted to specific joints (blue interaction boxes concentrated along one row). In contrast, our method effectively addresses this local bias using flexible grouping strategies. Fig. \ref{fig:3}c demonstrates the static topology learned through random initialization, which achieves better representation by overcoming local constraints.

Additionally, Tab. \ref{ex-ab5} includes results from experiments using distance matrices for initialization. However, all other experiments in this paper adopt random initialization for simplicity and generalization.

\begin{figure*}[h]
	\centering
		\centering
		\includegraphics[width=\linewidth]{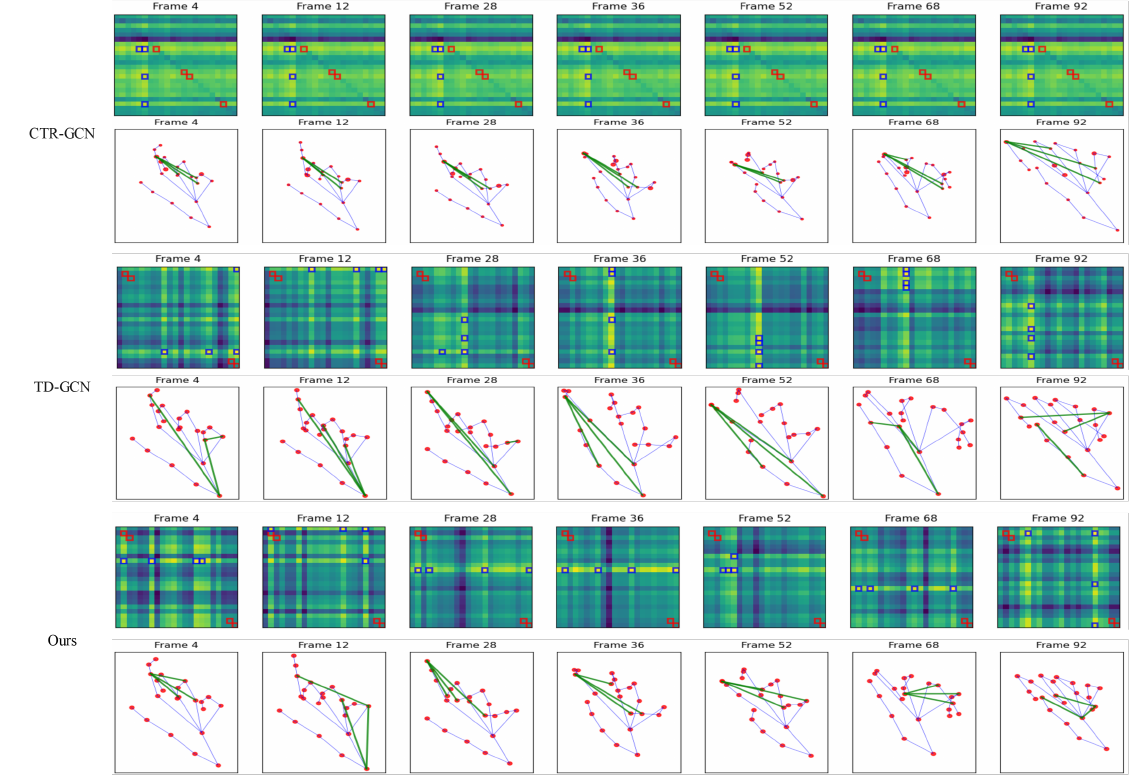}
	\caption{Visualization comparison of temporal-wise topology graphs (last layer) with CTR-GCN and TD-GCN. Gesture class : Grap. Blue boxes indicate the maximum four distant interactions, and red indicates the maximum four self interactions.}
	\label{fig:4}
\end{figure*}

\begin{figure}[h]
	\centering
		\centering
		\includegraphics[width=\linewidth]{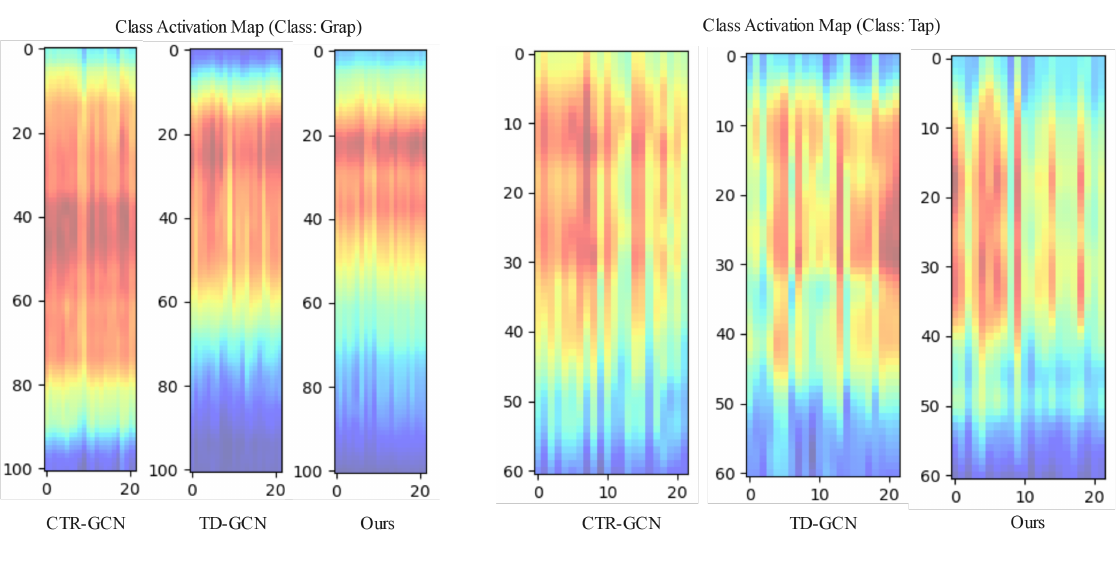}
	\caption{Class activation mapping results for action sample: Grap, Tap. The horizontal axis represents the joint index and the vertical axis represents the frame (temporal) index.}
	\label{fig:5}
\end{figure}

\begin{figure*}[h]
	\centering
		\centering
		\includegraphics[width=\linewidth]{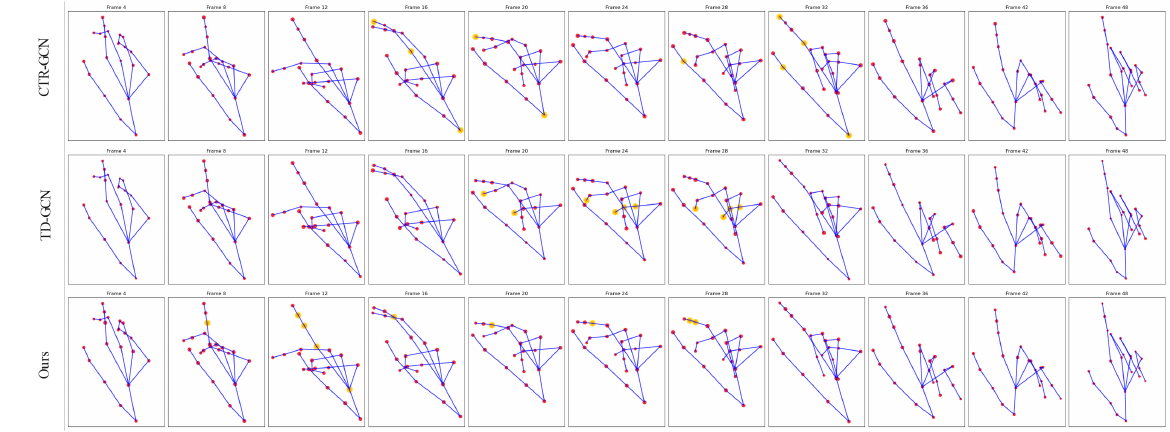}
	\caption{Visualization of the CAM results projected onto the skeletal gesture sequence (Tap). The yellow joints indicate the 10 most weighted joints in the CAM.}
	\label{fig:6}
\end{figure*}

\subsection{Comparisons with State-of-the-art Models}
We compare our DSTSA-GCN model with state-of-the-art methods on two gesture datasets, SHREC’17 Track and DHG-14/28, as well as two action datasets, NTU-RGB+D and NTU-RGB+D 120. The results are presented in Tab. \ref{ex:gesture_datasets} and Tab. \ref{ex:action datasets}. Following the standard practice, we report the fusion results for four data modalities: joint, bone, joint motion, and bone motion patterns. 

Additionally, we provide a qualitative comparison of temporal dynamic topology visualization and class activation mapping (CAM) with two other mainstream methods in Fig. \ref{fig:4}, Fig. \ref{fig:5}, and Fig. \ref{fig:6}.


\noindent \textbf{Gesture Recognition}.
As shown in Tab. \ref{ex:gesture_datasets}, on the SHREC’17 Track dataset, the DSTSA-GCN achieves the best performance (97.74\%) for 14 gesture classes and the second-best (95.37\%) for 28 gesture classes. Compared to CTR-GCN, which lacks temporal topology modeling, the accuracy improves by 1.64\% and 0.97\%, respectively. When compared to TD-GCN, which also includes temporal topology modeling, the improvements are 0.72\% and 0.01\%, respectively. 

On the DHG-14/28 dataset, which uses a leave-one-subject-out cross-validation strategy, we report all results in Tab. \ref{ex-dhg datasets}, using the final average accuracy for the final reported result in Tab. \ref{ex:gesture_datasets}. For both 14 and 28 gesture classes, our model achieves the best accuracies of 95.04\% and 93.57\%, respectively. These results represent improvements of 1.94\% and 3.07\% compared to CTR-GCN, and 1.14\% and 2.17\% compared to TD-GCN. 

Compared to CTR-GCN, our model can capture richer temporal topological information through the use of the GT-GC module. Furthermore, our model captures more distant joint interactions compared to TD-GCN (Fig. \ref{fig:3}c vs. Fig. \ref{fig:3}b). For STDA-GCN, while it uses a dynamic attention mechanism to update the adjacency matrix \( \mathbf{A} \) in GCN, addressing local bias to some extent, it still fixes the subset to 3 (as in ST-GCN, CTR-GCN, and TD-GCN), limiting the spatio-temporal topological expressiveness.


\noindent \textbf{Action Recognition}.
To assess the generalization ability of our model, we compare DSTSA-GCN with other state-of-the-art GCN methods on the NTU-RGB+D and NTU-RGB+D 120 datasets, as shown in Tab. \ref{ex:action datasets}. 

On the X-Sub and X-View benchmarks of the NTU-RGB+D dataset, the classification results after fusion achieve second-best accuracy of 92.78\% and best accuracy of 97.03\%. These results surpass CTR-GCN by 0.38\% and 0.23\%, respectively, and outperform SAN-GCN by 0.68\% and 0.83\%. 

On the X-Sub and X-Set benchmarks of the NTU-RGB+D 120 dataset, our model achieves third-best accuracy of 89.12\% and 90.97\%, surpassing CTR-GCN by 0.22\% and 0.37\%, respectively, and outperforms SAN-GCN by 0.43\% and 0.87\%.

It is important to note that while SAN-GCN also uses a grouping strategy to determine the optimal topology graphs, its topology is shared across all channels within each group, unlike the channel-specific topology used in CTR-GCN, which contributes to its lower effectiveness. Furthermore, SAN-GCN still lacks temporal-specific topology modeling capabilities. 

Moreover, it is noteworthy that Block-GCN shares our perspective on the static topology bias issue. It encodes weight matrices for the static topology using naturally connected topological distance matrices and Euclidean distance matrices. These weight matrices help enforce diversity in the static matrices.



\noindent \textbf{Temporal-wise Dynamic Topology and CAM Visualization}.
To validate the temporal-relative dynamic topology modeling capability, we extract frames \{4, 12, 28, 36, 52, 68, 92\} from the dynamic topology of the last layer of DSTSA-GCN (averaged over eight groups) in Fig. \ref{fig:4}. For the same sample (grap), we compare the results with the topologies of CTR-GCN and TD-GCN. In CTR-GCN, which lacks temporal topology modeling capability, the dynamic topology is shared across all frames. In contrast, both TD-GCN and DSTSA-GCN are able to capture distinct interaction topologies for different frames.

To more intuitively assess the ability of the model to perceive the spatio-temporal location of crucial joints, we apply Class Activation Mapping (CAM) \cite{zhou2016learning}, inspired by visualization techniques commonly used in image-related tasks. As shown in Fig. \ref{fig:5}, CAM highlights the most important spatio-temporal regions in the skeleton sequence. The sample \textit{Gap} has a length of 95, and the sample \textit{Tap} has a length of 59. 

For the \textit{Gap} sample, our model recognizes that all joints have similar importance, but it can distinguish frames where interaction information is more critical. In the \textit{Tap} sample, our model more accurately identifies the position of the index finger (index = 9, as defined in \cite{de20173d}) compared to TD-GCN. 

In Fig. \ref{fig:6}, we visualize the CAM values across frames \{4, 8, 12, 16, 20, 24, 28, 32, 36, 42, 48\} of sample \textit{Tap}. The joints' radius correspond to the value in the CAM diagram.


\section{Conclusion}
In this work, we propose DSTSA-GCN, a skeleton-based gesture recognition model that integrates both channel-specific and temporal-specific topology modeling to capture more expressive spatio-temporal features. By leveraging shared feature transformation functions (STCA) across both channel-wise and temporal-wise topology modeling, the model enhances its ability to sense variations in spatio-temporal locations. Furthermore, inspired by the concept of grouped convolution, we design a corresponding grouped graph convolution that maintains model complexity while mitigating local bias issues in the static topology at deeper layers. 

Through comprehensive experiments on two gesture datasets and two full-body action datasets, we not only demonstrate the ability of DSTSA-GCN to effectively extract flexible gesture features, but also validate its promising potential for full-body action recognition tasks.

\section*{Declaration of Competing Interest}
The authors declare that they have no known competing financial interests or personal relationships that could have appeared to influence the work reported in this paper.

\bibliographystyle{elsarticle-num} 
 \bibliography{mybib}

\end{document}